*Башмур К.А., Тынченко В.С., Бухтояров В.В., Сарамуд М.В.*

# ВЫБОР ТЕХНОЛОГИИ И ОЦЕНКА ПРОИЗВОДСТВЕННЫХ ВОЗМОЖНОСТЕЙ РОБОТА 3D-ПРИНТЕРА ДЛЯ СОЗДАНИЯ ЭЛЕМЕНТОВ ЭКСПЕРИМЕНТАЛЬНОГО ОБОРУДОВАНИЯ ДЛЯ ПРОИЗВОДСТВА КОМПОНЕНТОВ БИОТОПЛИВ

Сибирский федеральный университет, Красноярск

Bashmur K.A., Tynchenko V.S., Bukhtoyarov V.V., Saramud M.V.

# CHOICE OF TECHNOLOGY AND EVALUATION OF THE PRODUCTION CAPABILITIES OF A 3D PRINTER ROBOT FOR CREATING ELEMENTS OF EXPERIMENTAL EQUIPMENT FOR THE PRODUCTION OF BIOFUEL COMPONENTS

Siberian Federal University, Krasnoyarsk

Elements of experimental equipment for the production of biofuel components must meet high reliability and safety requirements. At the same time, in the course of research on the subject of creating equipment for the production of biofuels, a variable range of equipment is regularly proposed and should be checked. The manufacture of elements of such equipment by traditional methods is expensive and inefficient, time-consuming, which negatively affects the speed of scientific research. To this end, it is proposed to develop a robotic 3D printing complex that provides maximum flexibility in creating mock-ups and test samples of equipment for the production of biofuel components. The article discusses the experience of successfully creating equipment elements for the production of fuels using 3d printing. Next, the choice of a robotization scheme for a 3D printing installation is described and the choice of printing technology is substantiated. The article also presents the results of calculating the parameters of the 3v-printer robot and the results of calculating the similarity parameters for the implementation and evaluation of control algorithms. The results of a numerical experiment for calculating the strength characteristics of equipment elements manufactured using the selected 3d printing technology are presented.

Технологии традиционных методов изготовления деталей резанием достигли своего апогея, последним этапом развития стало числовое управление станков, чем удалось добиться большей точности и увеличить скорость изготовления деталей, однако развитие остановилось. Традиционные методы ремонта и восстановления способны вернуть полную работоспособность, но не надежность, следовательно, после каждого последующего ремонта наработка оборудования уменьшается, пока не наступит критический момент, когда ремонт нецелесообразен, и оборудование списывают в утиль.

За последние 10 лет технологии машиностроения шагнули далеко вперед. Появилось различное множество новых конструкционных материалов, а так же методов их обработки и нанесения, превосходящих традиционные конструкционные материалы по прочностным и массогабаритным характеристикам.

Одним из передовых аддитивных методов изготовления деталей является 3D-печать. Вместо традиционного метода обработки резанием, где с заготовки удаляется излишний материал, 3D-печать позволяет создавать деталь послойно, без траты лишних ресурсов и без получения отходов.

**Опыт создания экспериментального оборудования для производства компонентов биотоплив с использованием 3d-печати**

Анализ информации об использовании 3d-печати для производства элементов топливного технологического оборудования, в том числе реакторов, позволяет утверждать о возможности реализации предполагаемого подхода. Так, инженерами-конструкторами одной из компаний - производителей оборудования для предприятий топливно-химического цикла была разработана 3d-модель микрореактора, представленная на рисунке 1.

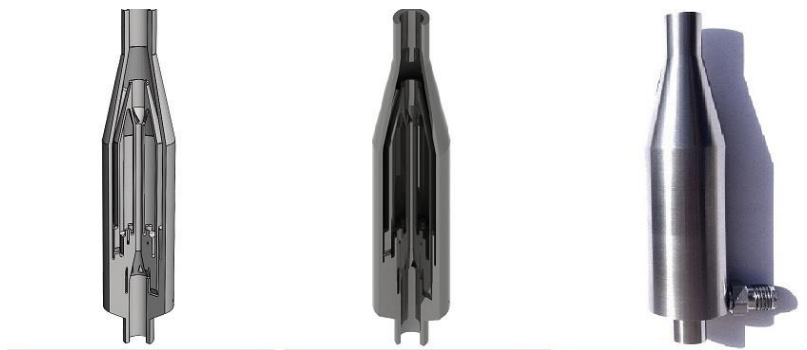

Рисунок 1 – Химический реактор

После проектирования модель микрореактора была напечатана из нержавеющей стали на 3d-принтере. После печати наружную поверхность микрореактора отполировали и нарезали резьбу для штуцера подачи водорода.

Серия испытаний проводилась в течение нескольких месяцев. Режим работы установок 24 часа в сутки. Давление в реакторе было от 30 до 50 бар, в зависимости от плана эксперимента, температура от 150ºС до 500ºС. В реакторе получали высокооктановые компоненты бензина, необходимые для производства бензинов класса Евро-5. После успешных испытаний заказчик решил заказать уже партию микрореакторов, напечатанных на 3D принтере. Стоимость микрореактора составила около 500 долларов, что для аналогичных микрореакторов, изготовленных традиционными способами, примерно в 10-15 раз дешевле. Процесс традиционного изготовления — это сварка, токарная обработка, сборная конструкция. Ориентировочная стоимость микрореактора 5000-7500$ за штуку.

**Роботизация 3d-печати**

Стационарный принтер в современных условиях при работе со сложными технологическими объектами, не обеспечивает должный уровень гибкости и свободы перемещения исполнительных механизмов и устройств.

Кроме того, повышенные требования к точности обуславливают возникающие требования к самообслуживанию, самонастройке и калибровке машины. Человеческий фактор при этом сводится к минимуму: роль человека сводится только лишь к тому, чтобы сформулировать задание и нажать на кнопку. Поэтому необходимо модернизировать не только конструкцию принтера, но и систему управления.

Авторами предлагается вместо стандартной рамы и шаговых двигателей использовать роботизированный манипулятор, устанавливаемый и перемещающийся в камере спекания. Таким образом, манипулятор будет иметь пять степеней свободы, а приводы современных роботов достаточно точны, для обеспечения точного позиционирования оборудования в пространстве.

**Технология работы робота 3d-принтера**

Проведя анализ существующих установок для аддитивного производства, можно сделать вывод, что предполагаемый к созданию робота 3d-принтер, работающий по технологии выборочной лазерной плавки, на сегодняшний день, является технологичным и обладающим необходимыми производственными возможностями. Это обусловлено большим количеством материалов изготовления деталей, мощными лазерами, наилучшей точностью и объемной камерой спекания, позволяющей изготавливать довольно габаритные детали. Поэтому все расчеты проводились именно для предлагаемого к созданию робота 3d-принтера.

В зависимости от требований к поверхности и точности изготовления, а так же материала изготовления, скорость печати проектируемого робота 3d-принтера варьируется от 10 см$^3$/час до 100 см$^3$/час. В таблице 1 приведены расчетные данные времени изготовления деталей при различных требованиях к поверхности и точности [20].

Таблица 1 – Расчет времени изготовления деталей

| Наименование детали | Объем, см$^3$ | А, час | Б, час | В, час |
|---|---|---|---|---|
| Подшипник скольжения | 2,9 | 0,4 | 0,3 | 0,2 |
| Центробежное колесо | 81 | 8 | 4,3 | 3,2 |
| Ротор турбоагрегата | 195 | 21 | 13,1 | 9,6 |
| Втулка резьбовая | 8,2 | 1,5 | 0,9 | 06 |
| А- 0,02мм; Б-0,05мм; В- 0,1мм | | | | |

**Расчет критериев подобия лазерного спекания**

Произведем расчет критериев подобия для робота 3d-принтера работающего по технологии выборочной лазерной плавки.

Система представляет собой оптическую систему (лазер) с координатным управлением, устройство для нанесения слоя порошкового материала и непосредственно сам порошок.

1. Выявим параметры, которые определяют процесс спекания металлического порошка:

$t_c$ = [с]; V= [мкм$^3$]; T= [ K]; z= [кДж/(кг*°С)]; М= [кг]; Е= [Вт]; ρ= [кг/м$^3$]

Функциональная зависимость, подлежащая исследованию:

$$F(t_c; V; T; z; M; E; ρ)=0 \qquad (1)$$

2. Выберем четыре независимые переменные применительно к системе *LMTΘ* (*L* – линейный размер, м; *M* – масса, кг; *T* – время, с; *Θ*- температура, К)

В качестве основных параметров примем:
$t_c$ = [с]; V= [мкм$^3$]; T= [ K]; z= [кДж/(кг*°С)].

3. Определим размерность каждого базисного параметра:

$$t_c = [L]^0[M]^0[T]^1[Θ]^0; \qquad (2)$$

$$V = [L]^3[M]^0[T]^0[\Theta]^0; \qquad (3)$$

$$T = [L]^0[M]^0[T]^0[\Theta]^1; \qquad (4)$$

$$z = [L]^2[M]^{-1}[T]^{-2}[\Theta]^{-1} \qquad (5)$$

Остальные параметры уравнения (1) примут вид:

$$\rho = [L]^{-3}[M]^1[T]^0[\Theta]^0; \qquad (6)$$

$$E = [L]^2[M]^0[T]^{-3}[\Theta]^0; \qquad (7)$$

$$M = [L]^0[M]^1[T]^0[\Theta]^0 \qquad (8)$$

4. Проверяем правильность сделанного выбора по числу базисных параметров, составим матрицу размерностей и вычислив ее определитель:

$$D0 := \begin{pmatrix} 0 & 0 & 1 & 0 \\ 3 & 0 & 0 & 0 \\ 0 & 0 & 0 & 1 \\ 2 & -1 & -2 & -1 \end{pmatrix}$$

$$|D0| = 3$$

Исходя из полученного значения определителя матрицы, можно сделать вывод, что $D0 \neq 0$, следовательно, значение базисных параметров и их число выбрано правильно и величины $t_c$, V, T, z действительно независимые.

5. Составим выражения для оставшихся критериев подобия. В общем виде, их можно записать в виде дробей:

$$\pi_5 = \frac{[\rho]}{[t_C]^{k1}[V]^{k2}[T]^{k3}[z]^{k4}};$$
(9)

$$\pi_6 = \frac{[E]}{[t_C]^{k5}[V]^{k6}[T]^{k7}[z]^{k8}};$$
(10)

$$\pi_7 = \frac{[M]}{[t_C]^{k9}[V]^{k10}[T]^{k11}[z]^{k12}};$$
(11)

Нахождение критериев подобия заключается в отыскании значений показателей степени.

6. Определим значения показателей степени k1-k12. Для этого найдем определители матриц размерностей для величин ρ, E, M заменив *i*-ую строку в матрице на строку составленную из показателей степени величин взятых из уравнений (6,7,8):

$$D1 := \begin{pmatrix} -3 & 1 & 0 & 0 \\ 3 & 0 & 0 & 0 \\ 0 & 0 & 0 & 1 \\ 2 & -1 & -2 & -1 \end{pmatrix} \qquad D2 := \begin{pmatrix} 0 & 0 & 1 & 0 \\ -3 & 1 & 0 & 0 \\ 0 & 0 & 0 & 1 \\ 2 & -1 & -2 & -1 \end{pmatrix} \qquad D3 := \begin{pmatrix} 0 & 0 & 1 & 0 \\ 3 & 0 & 0 & 0 \\ -3 & 1 & 0 & 0 \\ 2 & -1 & -2 & -1 \end{pmatrix}$$

$|D1| = -6$ $\qquad\qquad |D2| = -1 \qquad\qquad |D3| = -3$

$$D4 := \begin{pmatrix} 0 & 0 & 1 & 0 \\ 3 & 0 & 0 & 0 \\ 0 & 0 & 0 & 1 \\ -3 & 1 & 0 & 0 \end{pmatrix} \qquad D5 := \begin{pmatrix} 2 & 0 & -3 & 0 \\ 3 & 0 & 0 & 0 \\ 0 & 0 & 0 & 1 \\ 2 & -1 & -2 & -1 \end{pmatrix} \qquad D6 := \begin{pmatrix} 0 & 0 & 1 & 0 \\ 2 & 0 & -3 & 0 \\ 0 & 0 & 0 & 1 \\ 2 & -1 & -2 & -1 \end{pmatrix}$$

$|D4| = -3$ $\qquad\qquad |D5| = -9 \qquad\qquad |D6| = 2$

$$D7 := \begin{pmatrix} 0 & 0 & 1 & 0 \\ 3 & 0 & 0 & 0 \\ 2 & 0 & -3 & 0 \\ 2 & -1 & -2 & -1 \end{pmatrix} \quad D8 := \begin{pmatrix} 0 & 0 & 0 & 1 \\ 3 & 0 & 0 & 0 \\ 0 & 0 & 0 & 1 \\ 2 & 0 & -3 & 0 \end{pmatrix} \quad D9 := \begin{pmatrix} 0 & 1 & 0 & 0 \\ 3 & 0 & 0 & 0 \\ 0 & 0 & 0 & 1 \\ 2 & 0 & -3 & 0 \end{pmatrix}$$

$|D7| = 0$      $|D8| = 0$      $|D9| = -9$

$$D10 := \begin{pmatrix} 0 & 0 & 1 & 0 \\ 0 & 1 & 0 & 0 \\ 0 & 0 & 0 & 1 \\ 2 & 0 & -3 & 0 \end{pmatrix} \quad D11 := \begin{pmatrix} 0 & 0 & 1 & 0 \\ 3 & 0 & 0 & 0 \\ 0 & 1 & 0 & 0 \\ 2 & 0 & -3 & 0 \end{pmatrix} \quad D12 := \begin{pmatrix} 0 & 0 & 1 & 0 \\ 3 & 0 & 0 & 0 \\ 0 & 0 & 0 & 1 \\ 0 & 1 & 0 & 0 \end{pmatrix}$$

$|D10| = 2$      $|D11| = 0$      $|D12| = -3$

Теперь, зная определители матриц размерности элементов ρ, E, M найдем численные значения показателей степени k1-k11 по формуле (12):

$$ki = \frac{Di}{D0} \qquad (12)$$

Численные значения будут равны:

$k1 = \frac{D1}{D0} = \frac{-6}{3} = 2;\ k2 = \frac{D2}{D0} = \frac{-1}{3};\ k3 = \frac{D3}{D0} = \frac{-3}{3} = -1;\ k4 = \frac{D4}{D0} = \frac{-3}{3} = 1;$

$k5 = \frac{D5}{D0} = \frac{-9}{3} = -3;\ k6 = \frac{D6}{D0} = \frac{2}{3};\ k7 = \frac{D3}{D0} = 0;\ k8 = \frac{D8}{D0} = 0;$

$k9 = \frac{D9}{D0} = \frac{-9}{3} = -3;\ k10 = \frac{D10}{D0} = \frac{-2}{3};\ k11 = \frac{D3}{D0} = 0;\ k12 = \frac{D12}{D0} = \frac{-3}{3} = 1.$

7. Используя значения показателей и данные уравнений (9,10,11), получим окончательные значения критериев:

$$\pi_5 = \frac{[\rho]}{[t_C]^{-2}[V]^{-1/3}[T]^{-1}[z]^{-1}} = \rho \cdot t_C{}^2 \cdot \sqrt[3]{V} \cdot T \cdot z;$$

$$\pi_6 = \frac{[E]}{[t_C]^{-3}[V]^{2/3}[T]^0[z]^0} = \frac{E \cdot t_C{}^3}{\sqrt[3]{V^2}};$$

$$\pi_7 = \frac{[M]}{[t_C]^{-3}[V]^{-2/3}[T]^0[z]^{-1}} = M \cdot t_C{}^3 \cdot \sqrt[3]{V^2} \cdot z$$

Согласно второй теореме подобия, уравнение описывающее спекание порошка твердого сплава представляется функциональной зависимостью из критериев подобия:

$$\Phi(\pi_5; \pi_6; \pi_7) = 0$$

или

$$\Phi\left(\rho \cdot t_C{}^2 \cdot \sqrt[3]{V} \cdot T \cdot z;\ \frac{E \cdot t_C{}^3}{\sqrt[3]{V^2}};\ M \cdot t_C{}^3 \cdot \sqrt[3]{V^2} \cdot z\right) = 0 \qquad (13)$$

В результате проведенных расчетов, были получены критерии подобия и выявлена функциональная зависимость, позволяющая на основе данных натурного объекта строить модель, адекватно отражающую физические процессы, протекающие при реальном лазерном спекании металлических порошков различных видов.

### 3.3 Имитация нагрузок крыльчатки в SolidWorks

Для воспроизведения нагрузок, испытываемых крыльчаткой турбокомпрессора при максимально интенсивной работе, создана трехмерная модель крыльчатки. Далее с помощью программного комплекса SolidWorks и пакета Simulation Xpress были произведены два вида расчетов: расчет на прочность и расчет максимальных перемещений.

Оба расчета производились с заданной температурой крыльчатки равной 450°C. Расчеты проводились с двумя материалами: сталь 40х и ТНМ20.

На рисунке 2 показаны результаты расчета на прочность крыльчатки, изготовленной из стали 40х.

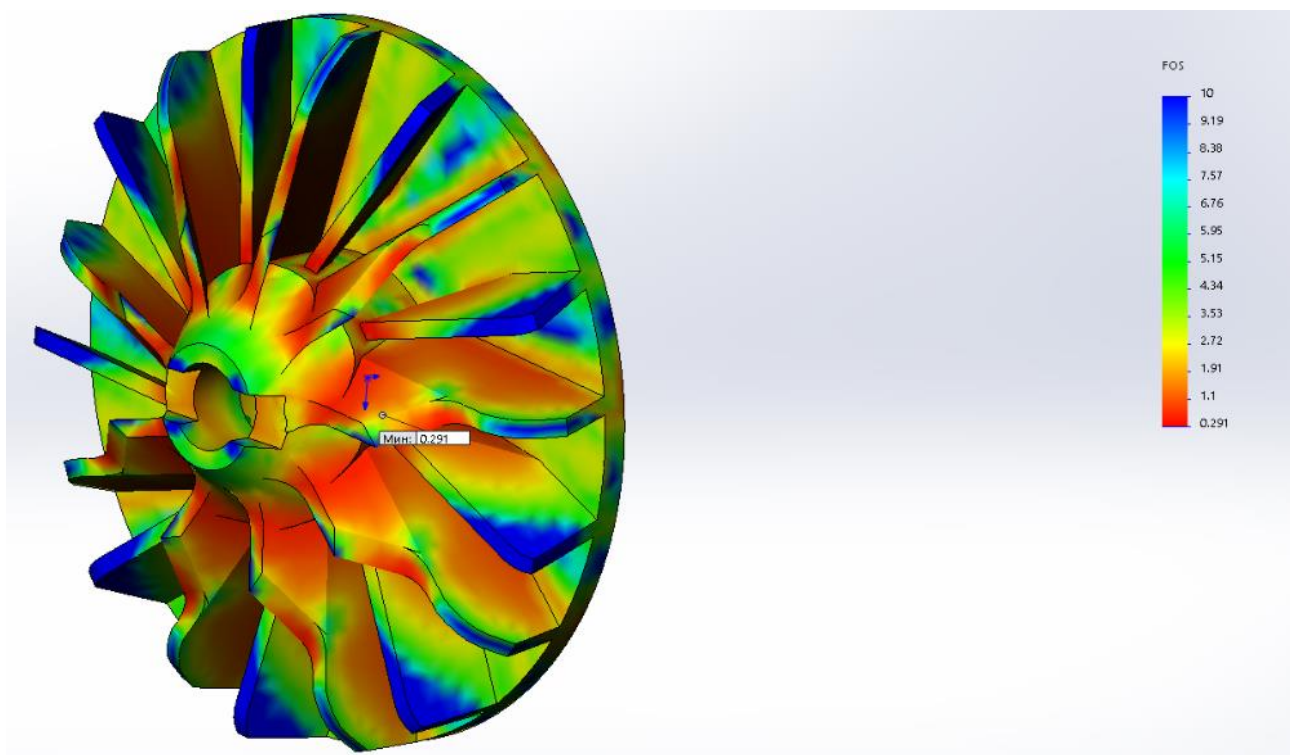

Рисунок 2 – Запас прочности (сталь 40Х)

На рисунке 3 приведены результаты расчета на прочность крыльчатки из THM20.

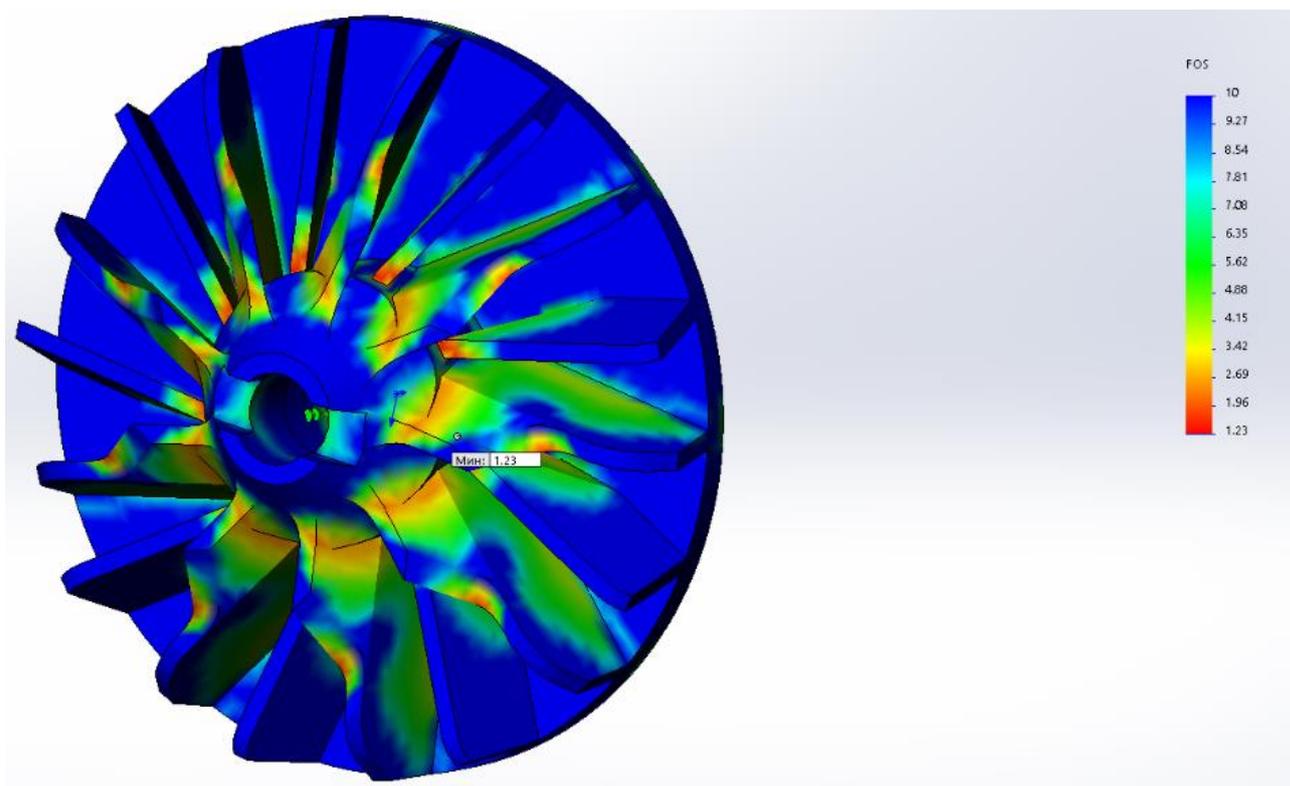

Рисунок 3 – Запас прочности (ТНМ20)

При первом расчете с материалом сталь 40Х минимальный коэффициент запаса прочности получился 0,291. При расчете с материалом ТНМ20 коэффициент запаса прочности 1,23. В итоге крыльчатка из ТНМ20 выигрывает в 4,2 раза по прочности, при идентичных размерах.

Вторым видом расчета является расчет на максимальные перемещения. Расчет крыльчатки из стали приведен на рисунке 4, максимальное перемещение составило 0,0787мм.

Расчет крыльчатки на максимальные перемещения, изготовленной из металлокерамики марки ТНМ20, показан на рисунке 5. Максимальные перемещения равны 0,0472, что в 1,67 раза меньше чем у стальной крыльчатки.

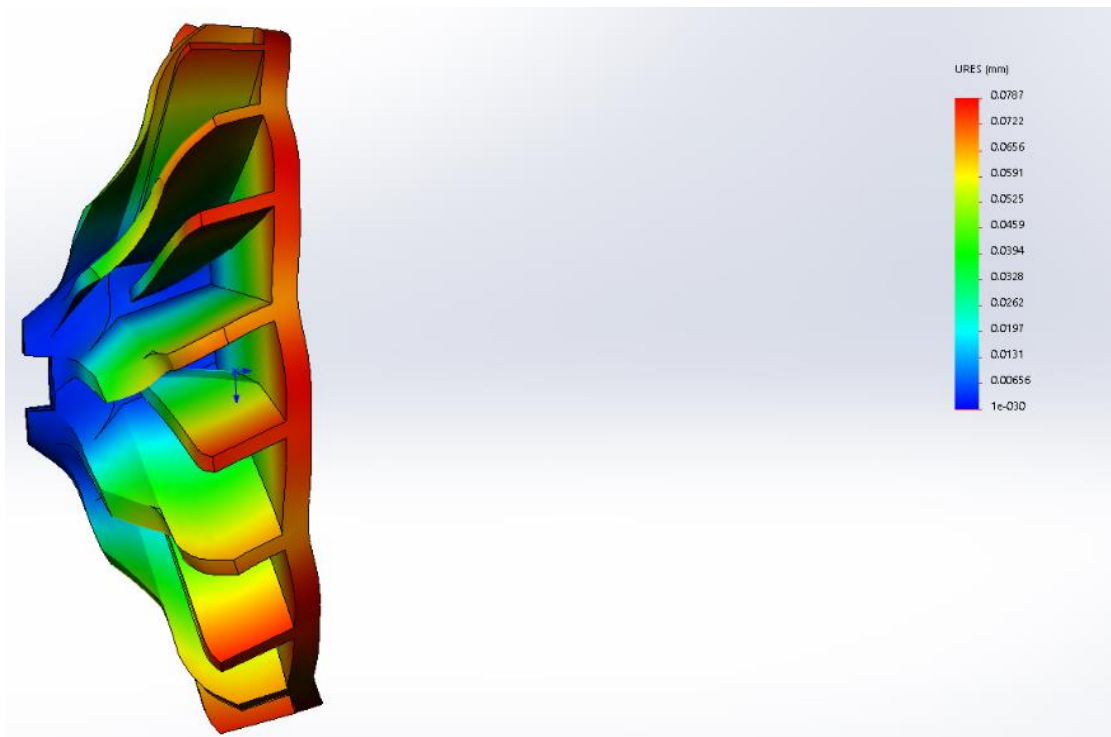

Рисунок 4 – Максимальные перемещения (сталь 40Х)

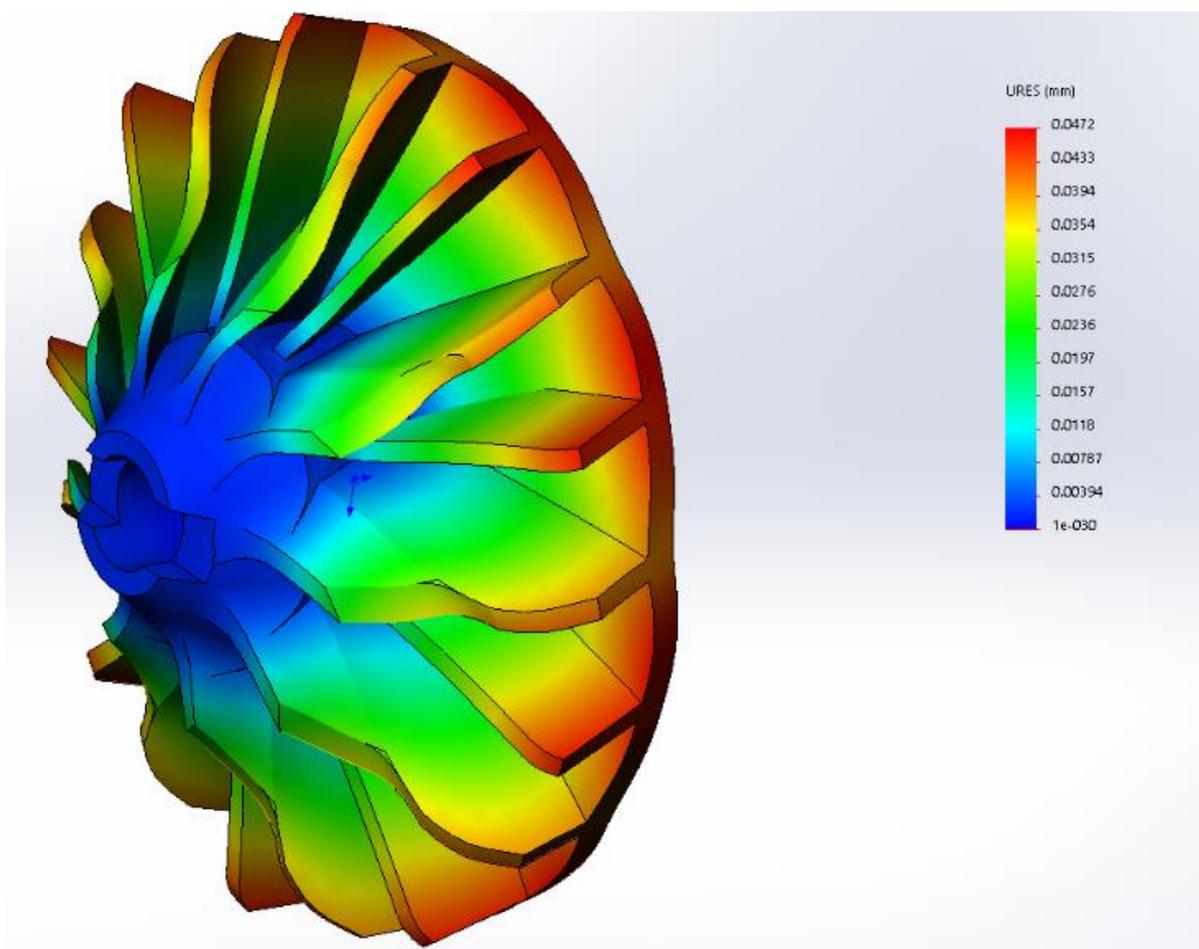

Рисунок 5 – Максимальные перемещения (ТНМ20)

**Выводы**

Робот 3d-принтер, использующий технологию выборочной лазерной плавки способен спекать любые металлические порошки, используемые в производстве. Размеры рабочей камеры робота 3d-принтера должны обеспечивать возможность изготовления крупногабаритных деталей, либо нескольких небольших, тем самым уменьшить время рабочего цикла. В ходе оценки возможностей робота 3d-принтера были рассчитаны критерии подобия спекания металлических порошков, позволяющие сократить время для настройки принтера под определенный материал. Полученная модель адекватно отражает физические процессы, протекающие при реальном лазерном спекании металлических порошков различных видов. Произведен прочностной расчет и расчет на максимальные перемещения крыльчатки турбокомпрессора, изготовленной из THM20, и заводской крыльчатки. Проведено сравнение полученных данных. Запас прочности у крыльчатки из THM20 в 4,2 раза выше, а максимальные перемещения в 1,67 раза меньше, чем у стандартной стальной крыльчатки. Это, наряду с положительным опытом зарубежных коллег позволяет утверждать о рациональности создания робота 3d-принтера для изготовления элементов экспериментального оборудования для производства компонентов биотоплив на основе технологии выборочной лазерной плавки.

**Дополнительные сведения**

созданной в рамках деятельности Научно-образовательного центра «Енисейская Сибирь».

## СПИСОК ЛИТЕРАТУРЫ